%% file: main.tex
\documentclass[10pt,twocolumn,letterpaper]{article}

\usepackage[pagenumbers]{cvpr} 

\usepackage{graphicx}
\usepackage{amsmath}
\usepackage{amssymb}
\usepackage{booktabs}
\usepackage{float}
\usepackage{bm}
\usepackage{bbm}
\usepackage{verbatim}
\usepackage{multirow}
\usepackage[pagebackref,breaklinks,colorlinks,citecolor=blue,linkcolor=blue,urlcolor=blue]{hyperref}
\usepackage{paralist}

\usepackage[capitalize]{cleveref}
\crefname{section}{Sec.}{Secs.}
\Crefname{section}{Section}{Sections}
\Crefname{table}{Table}{Tables}
\crefname{table}{Tab.}{Tabs.}

\def\cvprPaperID{0000} 
\def\confName{CVPR}
\def\confYear{2022}

\begin{document}

\title{Hierarchical Modular Network for Video Captioning \vspace{-2mm}}


\author{Hanhua Ye$^{1}$ \quad Guorong Li$^{1}$\thanks{Corresponding author.} \quad Yuankai Qi$^{2}$ \quad Shuhui Wang$^{3,4}$ \\ Qingming Huang$^{1,3,4}$ \quad Ming-Hsuan Yang$^{5}$ \\
$^{1}$University of Chinese Academy of Science, Beijing, China \\
$^{2}$Australian Institute for Machine Learning, The University of Adelaide, \\
$^{3}$Key Lab of Intell. Info. Process., Inst. of Comput. Tech., CAS, Beijing, China, \\
$^{4}$Peng Cheng Laboratory, Shenzhen, China, $^{5}$University of California, Merced \\
{\tt\small \{yehanhua20, liguorong\}@mails.ucas.ac.cn}, {\tt\small qykshr@gmail.com}, 
{\tt\small wangshuhui@ict.ac.cn}, \\{\tt\small qmhuang@ucas.ac.cn}, {\tt\small mhyang@ucmerced.edu}
}

\maketitle

\input{section/abs_intro}

\input{section/related}

\input{section/method}

\input{section/exper}

\input{section/limitation}

\input{section/conclu}

{\small
\bibliographystyle{ieee_fullname}
\bibliography{main}
}

\end{document}

%% file: section/abs_intro.tex
\begin{abstract}
Video captioning aims to generate natural language descriptions according to the content, where representation learning plays a crucial role. 
Existing methods are mainly developed within the supervised learning framework via word-by-word comparison of the generated caption against the ground-truth text without fully exploiting linguistic semantics. 
In this work, we propose a hierarchical modular network to bridge video representations and linguistic semantics 
from three levels before generating captions.
In particular, the hierarchy is composed of: (I) Entity level, which highlights objects that are most likely to be mentioned in captions. (II) Predicate level, which learns the actions conditioned on highlighted objects and is supervised by the predicate in captions. 
(III) Sentence level, which learns the global semantic representation and is supervised by the whole caption.
Each level is implemented by one module.
Extensive experimental results show that the proposed method performs favorably against the state-of-the-art models on the two widely-used benchmarks: MSVD 104.0\% and MSR-VTT 51.5\% in CIDEr score.
Code will be made available at \href{https://github.com/MarcusNerva/HMN}{https://github.com/MarcusNerva/HMN}. 

\end{abstract}


\vspace{-4mm}
\section{Introduction}
\begin{figure}[tb]
    \centering
    \includegraphics[width=.95\linewidth, scale=0.7]{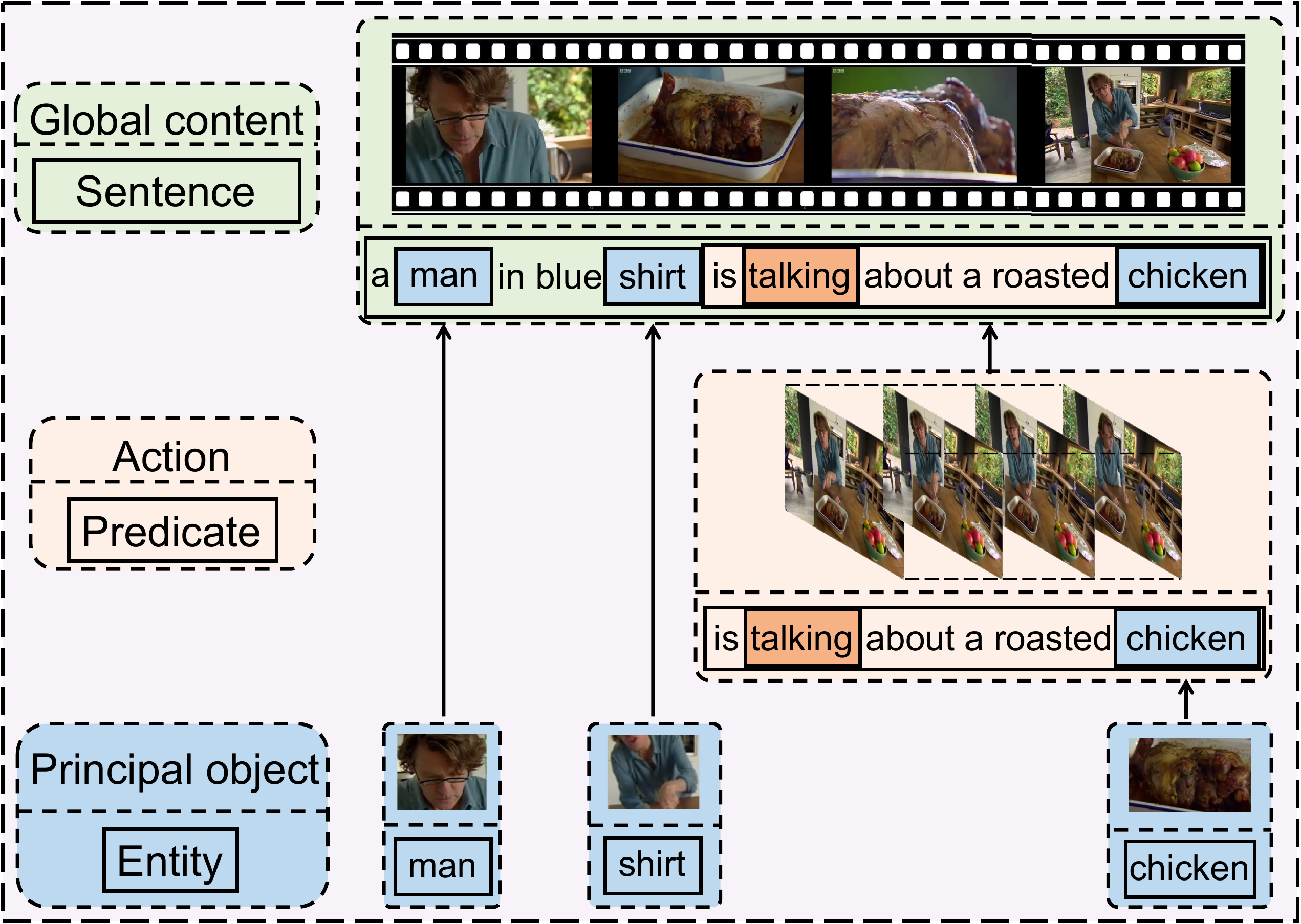}
    \caption{
    To effectively bridge video content and linguistic caption, we propose to supervise  video representation learning in a three-level hierarchical structure, i.e., the entity level, the predicate level, and the sentence level. 
    }
    \label{motivation}
     \vspace{-6mm}
\end{figure}

Video captioning aims to automatically generate natural language descriptions from videos, which plays an important role in numerous applications, such as assisting visually-impaired people, human-computer interaction, and video retrieval~\cite{DBLP:conf/iccv/VenugopalanRDMD15,DBLP:conf/naacl/VenugopalanXDRM15,DBLP:conf/ijcai/SongGGLZS17,DBLP:conf/cvpr/Wang00018,DBLP:conf/eccv/ChenWZH18,DBLP:conf/iccv/YaoTCBPLC15,DBLP:conf/cvpr/PanXYWZ16}.
Despite recent advances in this field, 
it remains a challenging task as a video usually contains rich and diverse content, but only some information is relevant to caption (\eg, two or three out of many objects are captured in a caption).

Existing methods aim to learn effective video representations to generate captions via recurrent decoders, which can be broadly categorized into two lines of work.  
The first one focuses on designing complex video encoders to learn better video representations~\cite{MGRMP,STG-KD,ORG-TRL,POS+CG,GRU-EVE,OA-BTG,DBLP:conf/aaai/ChenJ19}. 
For example, STG-KD~\cite{STG-KD} and ORG-TRL~\cite{ORG-TRL} build object relation graphs to reason the spatial and temporal relations between video objects. 
While GRU-EVE~\cite{GRU-EVE} applies Short Fourier Transform\cite{fourier} to embed temporal dynamics in visual features, POS+CG~\cite{POS+CG} develops a cross gating block to fuse appearance and motion features and make a comprehensive representation.
However, the optimization objectives of these methods are computed word-by-word as captions generating, disregarding the relevance between video representations and their linguistic counterpart. The other one focuses on narrowing the semantic gap between video representation and linguistic captions ahead of generating captions\cite{DBLP:conf/cvpr/PanMYLR16,DBLP:conf/cvpr/ShenLSLCJX17,SAAT}. 
For example, Pan~\etal~\cite{DBLP:conf/cvpr/PanMYLR16} learn to align the global representation of a video to the embedding of a whole caption.
In contrast, Shen~\etal~\cite{DBLP:conf/cvpr/ShenLSLCJX17} and Zheng~\etal~\cite{SAAT} associate  \textit{nouns} and \textit{verbs} with visual features to explore the video-language correspondence on a fine-grained level. 
These approaches are able to generate more accurate captions as more representative video embedding is learned. 
However, they either focus on global sentence correspondence or local word correspondence, which  disregard fine-grained details or global relevance.

In this work, we propose a hierarchical modular network to address the issues mentioned above. 
Our model aims to learn three kinds of video representations supervised by language semantics at different hierarchical levels as shown in Figure~\ref{motivation}:  (I) Entity level, which highlights objects that are most likely to be mentioned in captions and is supervised by \textit{entities}\footnote{Note that \textit{entities} are different from \textit{nouns}. \textit{Nouns} contain \textit{abstract nouns}, such as \textit{happiness and hunger.}, while \textit{entities} consist of object names, such as \textit{onion} and \textit{car}.} in the caption.  (II) Predicate level, which learns the actions conditioned on highlighted objects and is supervised by the predicate in the caption. (III) Sentence level, which learns the global
video representation supervised by the whole caption. 
Each level is implemented by one module. 
The motivation of our design is that objects usually serve as the cornerstone of a video caption, which can be the subject or object of an action as well as modifiers of subject and/or object.
Instead of learning the visual representation of {\textit{verbs}} along, we propose to learn video representation of {\textit{predicates}} (\textit{verb}+\textit{noun}). 
This helps reduce the correspondence errors from a multi-meaning  \textit{verb} to a specific video action embedding, such as the \textit{play} in \textit{playing soccer} and \textit{playing the piano}.
The global video content embedding supervised by the embedding of a whole caption enables the generated caption to have a reasonable meaning. 

It is worth noting that we propose a novel entity module. 
This module takes all the pre-extracted objects of a video as input and outputs a small set of principal objects that are most likely mentioned in a caption.
Motivated by the success of DETR~\cite{DETR} for object detection, our entity module is designed with a transformer encoder-decoder architecture. %
In contrast to DETR, our queries are enhanced by video content and supervised by entities in captions, which enables the model to select principal objects according to video scenarios. 

The contributions of this paper are summarized below:
\begin{compactitem}
\item We propose a hierarchical modular framework to learn multi-level visual representations at different granularity by associating them with their linguistic counterparts: the \textit{entity}, \textit{predicate}, and \textit{sentence}.
\item We propose a transformer-based entity module to learn to select principal objects that are most likely to be mentioned in captions. 
\item Our method performs favorably against the state-of-the-art models on two widely-used benchmarks: MSVD~\cite{DBLP:conf/acl/ChenD11} and MSR-VTT~\cite{DBLP:conf/cvpr/XuMYR16}.
\end{compactitem}

%% file: section/related.tex
\section{Related Work}

\noindent \textbf{From Template-based to CNN-based Methods.} There are a number of methods have been proposed for the video captioning task. 
Kojima~\etal~\cite{DBLP:journals/ijcv/KojimaTF02} and Krishnamoorthy~\etal~\cite{DBLP:conf/aaai/KrishnamoorthyMMSG13} propose to first generate words for objects and actions, and then fit predicted words into predefined sentence templates to generate captions. 
However, template-based methods are hard to generate flexible descriptions.
Inspired by the success of RNN and CNN, the encoder-decoder structure is widely utilized to generate descriptions with flexible syntactic structure\cite{DBLP:conf/naacl/VenugopalanXDRM15,DBLP:conf/cvpr/Wang00018,DBLP:conf/cvpr/PanXYWZ16}. 
In~\cite{DBLP:conf/iccv/VenugopalanRDMD15},  Venugopalan~\etal learn video representations by performing mean-pooling over CNN features of each frame, and exploit an LSTM~\cite{DBLP:journals/neco/HochreiterS97} to generate captions.
On the other hand, Yao~\etal~\cite{DBLP:conf/iccv/YaoTCBPLC15} design a temporal attention mechanism to aggregate relevant video segments given the state of text-generating RNN, modeling the global temporal structure of videos.
In addition to the most frequently used image and motion features, Hori~\etal~\cite{DBLP:conf/iccv/HoriHLZHHMS17} and Xu~\etal~\cite{DBLP:conf/mm/XuYZM17} exploit audio features to enrich the video representation. 
In~\cite{DBLP:conf/eccv/ChenWZH18}, Chen~\etal~ propose a PickNet to select informative frames from the video, removing redundant visual information. 
Recently, Wang~\etal~\cite{DBLP:conf/cvpr/Wang000T18} and Pei~\etal~\cite{DBLP:conf/cvpr/PeiZWKST19}  enhance the captioning quality by designing a memory network to organize multiple visual features. 

\noindent \textbf{Utilizing Detected Objects.} 
Objects  play an important role in generating captions, which are usually extracted  via pre-trained object detectors (\eg YOLO9000\cite{DBLP:conf/cvpr/RedmonF17}, Faster-RCNN\cite{Faster-R-CNN}, Mask RCNN\cite{DBLP:conf/iccv/HeGDG17}).
Significant efforts have been made to use object information of videos for captions. 
In~\cite{OA-BTG}, Zhang~\etal utilize a GRU~\cite{DBLP:conf/emnlp/ChoMGBBSB14} to capture object dynamic information from temporal trajectories.
Aafaq~\etal~\cite{GRU-EVE} exploit object labels (predicted by object detectors) to enhance the semantics of visual representation. 
On the other hand, Zheng~\etal~\cite{SAAT} adopt an dot-product attention mechanism to help determine the interactions between objects. 
In addition, Pan~\etal~\cite{STG-KD} and Zhang~\etal~\cite{ORG-TRL} employ graph convolutional networks~\cite{DBLP:conf/iclr/KipfW17} to perform relational reasoning among detected objects to enhance object-level representation.
These approaches can generate more accurate captions as detailed video information is mined.
However, all detected objects are used in these methods.
As usually only a small set of objects are mentioned in captions, the large number of redundant objects may negatively affect the captioning performance. 
In contrast, we propose an entity module to highlight principal objects that are most likely to be mentioned in the caption in this work, reducing the noise brought by redundant objects.

\begin{figure*}[!t]
    \centering
    \includegraphics[width=.9\linewidth, scale=0.7]{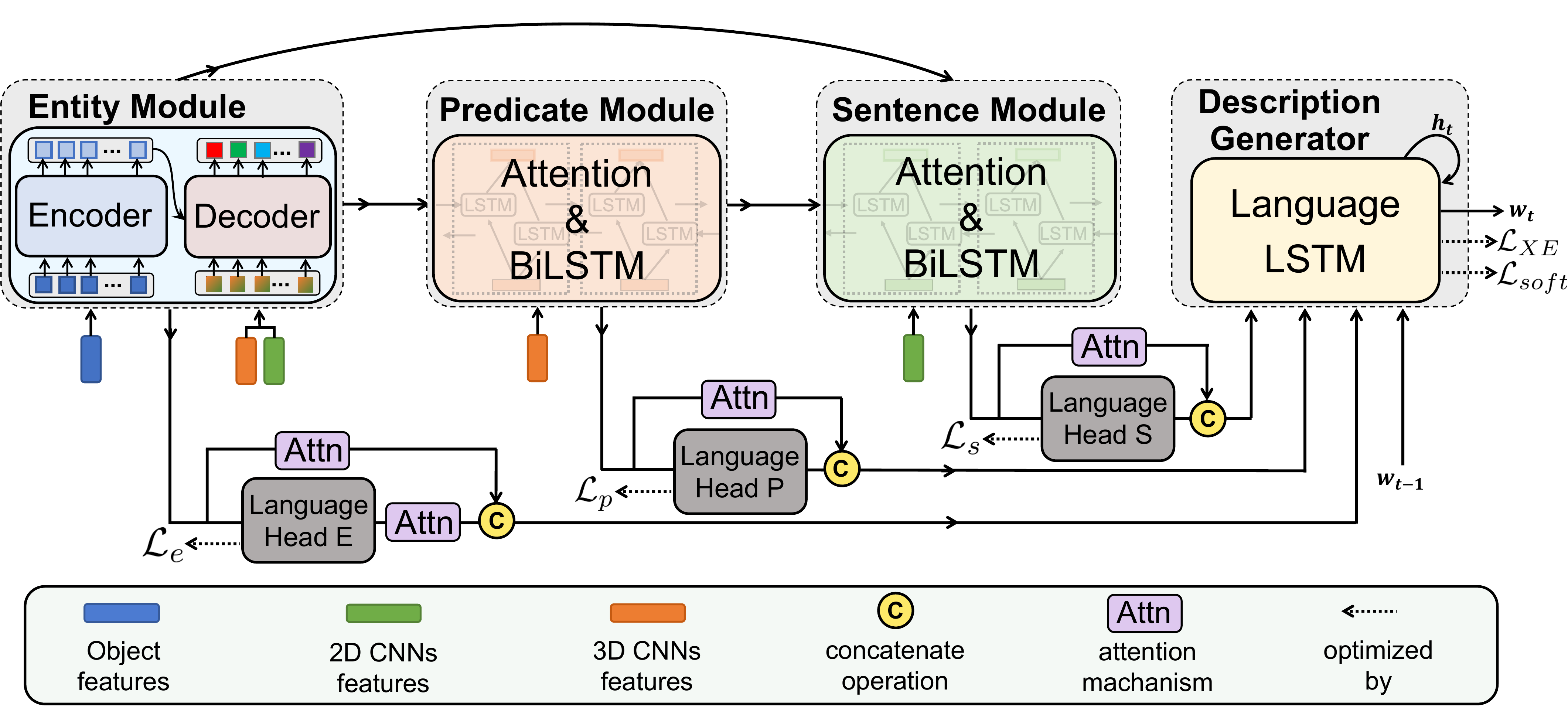}
    \caption{
    The proposed Hierarchical Modular Network serves as a strong video encoder, which bridges video representations and linguistic semantics from three levels via the entity (Sec.~\ref{sec:entity}), predicate  (Sec.~\ref{sec:predicate}), and sentence  (Sec.~\ref{sec:sentence}) modules. 
    Each module has its own input and linguistic supervision extracted from captions. 
    }
    \label{total}
    \vspace{-4mm}
\end{figure*}

\noindent \textbf{Transformer in Vision.} The success of transformer models~\cite{DBLP:conf/nips/VaswaniSPUJGKP17,DBLP:conf/naacl/DevlinCLT19,DBLP:journals/corr/abs-1907-11692} in natural language processing has attracted much interest in the computer vision community. 
Several methods have shown the effectiveness of transformers for vision tasks, such as image classification~\cite{DBLP:conf/iclr/DosovitskiyB0WZ21,DBLP:conf/icml/TouvronCDMSJ21,liu2021Swin}, object detection~\cite{DETR,DBLP:conf/iclr/ZhuSLLWD21}, video understanding\cite{DBLP:conf/iccv/SunMV0S19,DBLP:conf/cvpr/GirdharCDZ19}, and semantic segmentation~\cite{DBLP:conf/cvpr/ZhengLZZLWFFXT021}. 
Motivated by DETR~\cite{DETR}, which adaptively learn queries to detect objects,  in this work we develop a transformer-based entity module to highlight principal objects out of a large number of candidates.
Experiments and ablation studies show the effectiveness of this design.

%% file: section/method.tex
\section{Method}

As shown in Figure~\ref{total}, our  model follows the conventional encoder-decoder paradigm, where our Hierarchical Modular Network (HMN) serves as the encoder. 
Our HMN consists of the entity, predicate, and sentence modules. 
Equipped with language heads, these modules are designed to bridge video representations and linguistic semantics from three levels.
Our model operates as follows. 
First, taking all detected objects as input, the entity module outputs the features
of principal objects.
The predicate module encodes actions by combining features of principal objects and the video motion. 
Next, the sentence module encodes a global representation for the entire video content considering the global context  and features of previously generated objects and actions.
Finally, all features are concatenated  together and fed into the decoder to generate captions.

\subsection{Entity Module}
\label{sec:entity}

\begin{figure}[!t]
    \centering
    \includegraphics[width=0.95\linewidth, scale=0.6]{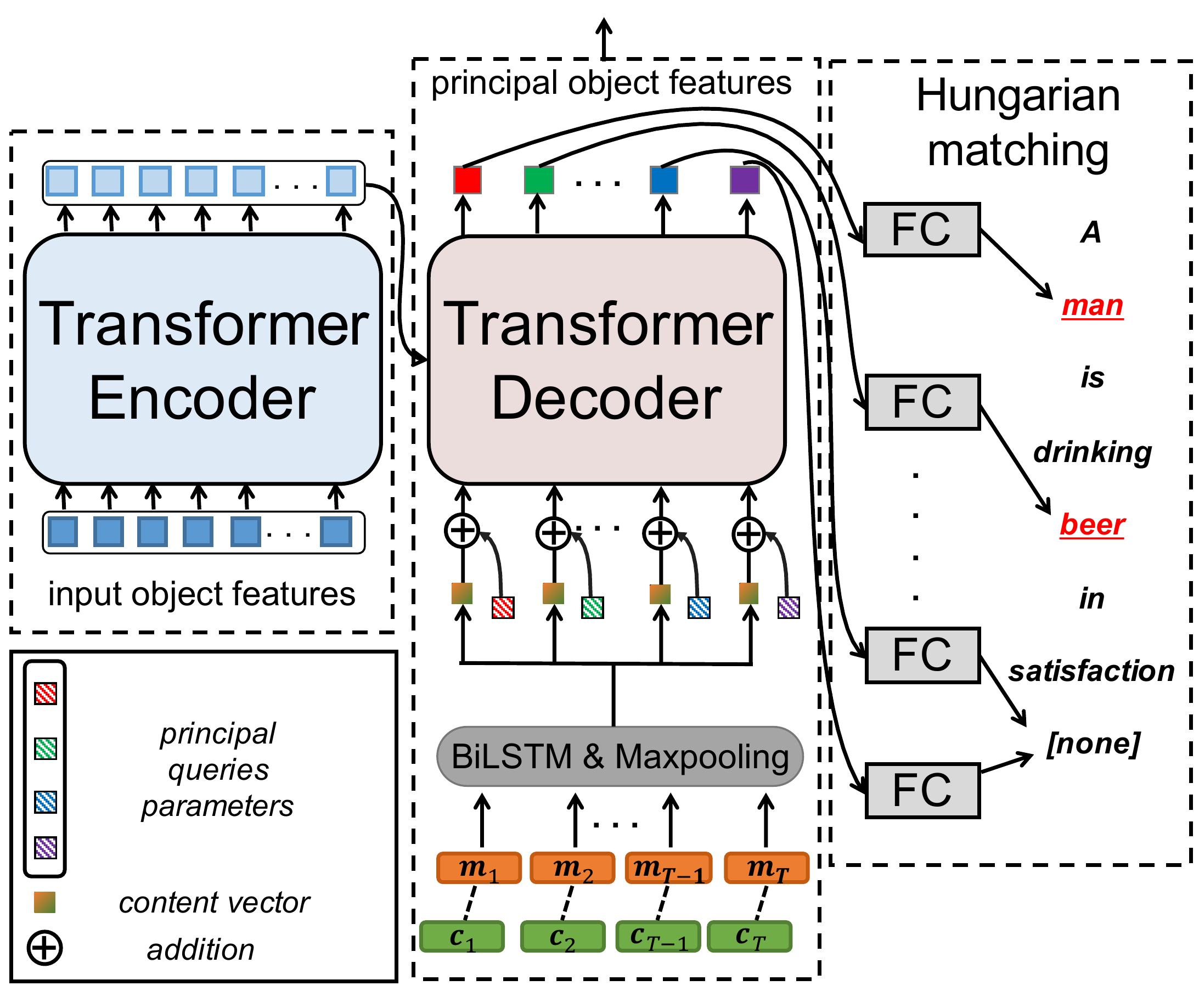}
    \caption{Main architecture of our entity module.}
    \label{object}
    \vspace{-4mm}
\end{figure}

Given a sequence of video frames, 
we uniformly select $T$ frames as keyframes and collect short-range video frames around keyframes as 3D cuboids. 
We exploit a pre-trained object detector\cite{Faster-R-CNN} to capture object regions from each keyframe, and cluster these regions according to  appearance and Intersection over Union (IoU)  among   bounding boxes. 
Then we apply the mean-pooling operation to these clusters to obtain the initial object features $\mathcal{O} = \{\bm{o}_{i}\}_{i=1}^{L}, \bm{o}_{i}\in \mathbb{R}^{ d_{o}}$, where $L$ and $d_{o}$ denote the number of video objects and size of object features.
As there are a large number of objects in a video, but only some are mentioned in captions, we design the entity module to learn to highlight these principal objects adaptively.

\noindent \textbf{Architecture.} 
Figure~\ref{object} illustrates the main architecture of our entity module, which consists of a transformer encoder and transformer decoder. 
This design is motivated by DETR~\cite{DETR}, which utilizes a transformer encoder-decoder architecture to learn a fixed set of object queries to directly predict object bounding boxes for the object detection task. 
Instead of simply detecting objects, we aim to determine the important ones in the video.
Due to significant difference between these two tasks, a simple application of DETR architecture performs poorly (see Section~\ref{sec:ablation}). 
As the key of DETR  is to learn queries, we design our own queries as detailed later.

The encoder maps input objects $\mathcal{O}$ to a set of representations:
\begin{equation}
    \mathcal{O}'=\{\bm{o}'_i\}_{i=1}^{L} = \bm{{\rm TransEncoder}}(\mathcal{O}),
\end{equation}
where $\bm{o}'_i \in \mathbb{R}^{  d_{model}}$.
Note that unlike conventional transformers, we discard  positional encodings of objects because the object spatial location is meaningless after we apply mean-pooling on object features. 

To enable the decoder to highlight principal objects, our decoder takes three types of inputs:
\begin{equation} \label{eq2}
    \mathcal{E} =\{\bm{e}_{i}\}_{i=1}^{N}= \bm{{\rm TransDecoder}}(\mathcal{O}', \mathcal{Q}, \{\bm{v}\}_{\times N}),
\end{equation}
where $\mathcal{O}'$ is the encoder outputs, $\mathcal{Q} = \{\bm{q}_{i}\}_{i=1}^{N}$  is randomly initialized $N$ parameters for queries. $\{\bm{v}\}_{\times N}$ are $N$ identical video content vectors to put each query $\bm{q}_{i}$ into a proper context and thus to facilitate decoding $N$ video-specific   principal object features.
To construct the video content vector $\bm{v}$, we first utilize the pre-trained 2D CNNs\cite{IRV2} and pre-trained 3D CNNs\cite{DBLP:conf/cvpr/HaraKS18} to extract the context features $\mathcal{C} = \{\bm{c}_{i}\}_{i=1}^{T}$ from keyframes, and motion features $\mathcal{M} = \{\bm{m}_{i}\}_{i=1}^{T}$ from 3D cuboids.
Then, $\mathcal{C}$ and $\mathcal{M}$ are concatenated and fed into a BiLSTM to generate a set of hidden states
$
    \mathcal{H} =\{\bm{h}_i\}_{i=1}^{T} = \bm{{\rm{BiLSTM}}}(\{[\bm{c}_i;\bm{m}_i]\}_{i=1}^T).
$
Next, we aggregate $\mathcal{H}$ by max-pooling to obtain an embedding $\bm{v} = \bm{{\rm{maxpool}}}(\mathcal{H})$ which represents video content. 

As the decoder output, principal object features $\mathcal{E}$ are later fed to the predicate and sentence modules, as well as the language head as shown in Figure~\ref{total}.
%
The  language head projects $\mathcal{E}$ to linguistic semantic space via a fully-connected layer
\begin{equation}
   \bar{\mathcal{E}}=\{\bar{\bm{e}}_i\}_{i=1}^{N} = \bm{{\rm FC}}(\mathcal{E}),
\end{equation}
where $\bar{\bm{e}}_i \in \mathbb{R}^{d_s}$. This language head is supervised by embeddings of entities from captions as detailed in the next sub-section.

\noindent \textbf{Loss function.}  
We exploit the \textit{entities} in captions to supervise our entity module. 
Specifically,
we first obtain the ``synonymy label" in \textit{WordNet}\footnote{https://wordnet.princeton.edu} of each \textit{noun} in ground-truth captions. 
And then, we choose \textit{nouns} with labels ``object.n.01", ``causal\_agent.n.01", and ``matter.n.03" as \textit{entities}, discarding \textit{abstract nouns}.

Assume that $M$ entities are extracted from a caption, we employ the pre-trained SBERT~\cite{SBert} as text encoder to compute the entities embeddings $\mathcal{N} = \{\bm{n}_i\}_{i=1}^{M}$, where $\bm{n}_i \in \mathbb{R}^{ d_s}$. We set $N$ larger than the typical number of \textit{entities} in captions, and pad $\mathcal{N}$ to size $N$ with $\varnothing$ (no \textit{entity}).
Then, we search for an optimal assignment $\hat{\sigma}$ between $\mathcal{N}$ and $\Bar{\mathcal{E}}$ with minimum distance cost:
\begin{equation}
    \hat{\sigma} = \mathop{\arg\min}_{\sigma \in \Omega_{N}}\sum_{i}^{N}Dist(\bm{n}_{i}, \bar{\bm{e}}_{\sigma(i)}), 
\end{equation}
where the $\Omega_{N}$ is the assignment search space for $N$ elements, and $Dist(\bm{n}_{i}, \bar{\bm{e}}_{\sigma(i)})$ is a pair-wise matching cost between the entity embedding $\bm{n}_{i}$ and predicted linguistic embedding  $\bar{\bm{e}}_{\sigma(i)}$ by
\begin{equation}
    Dist(\bm{n}_{i}, \bar{\bm{e}}_{\sigma(i)})  = \mathbbm{1}_{\{\bm{n}_{i} \neq \varnothing\}} \cdot (1 - \frac{\bm{n}_{i} \cdot  \bar{\bm{e}}_{\sigma(i)}}{|\bm{n}_{i}| \cdot |\bar{\bm{e}}_{\sigma(i)}|}).  
\end{equation}
This optimal assignment is computed efficiently with Hungarian algorithm~\cite{kuhn1955hungarian} as in DETR~\cite{DETR}.
Finally, according to the optimal assignment $\hat{\sigma}$, we optimize our entity module by minimizing the distance between $\mathcal{N}$ and $\Bar{\mathcal{E}}$:
\begin{equation}
  \mathcal{L}_{e} = \sum_{i}^{N}Dist(\bm{n}_{i}, \bar{\bm{e}}_{\hat{\sigma}(i)}). 
\end{equation}

\subsection{Predicate Module} 
\label{sec:predicate}

Our predicate module is designed to learn action representations that bridge the video action information and the \textit{predicate} of the caption.
As a \textit{predicate} usually consists of a \textit{verb} and its recipient, our predicate module concatenates the initial motion features $\mathcal{M} = \{\bm{m}_i\}_{i=1}^{T}$ and motion-related object features $\mathcal{M}^{e} = \{\bm{m}_{i}^{e}\}_{i=1}^{T}$ as input. For each motion $\bm{m}_i$, we compute its motion-related object features $\bm{m}_{i}^{e}$  by attentively summarizing principal object features $\mathcal{E}$  via
\begin{align}
\bm{m}_{i}^{e}  = \sum_{k=1}^{N}& \alpha_{i,k}\bm{e}_{k}, \nonumber  \\
\alpha_{i,k} = {\rm exp}(\hat{\alpha}_{i,k}) /& \sum_{j=1}^{N}{\rm exp}(\hat{\alpha}_{i, j}), \\
 \hat{\alpha}_{i,k} = \bm{{\rm w}}_{a}^{\top}{\rm tanh}(\bm{{\rm W}}_{a}^{\top}&\bm{m}_{i}  + \bm{{\rm U}}_{a}^{\top}\bm{e}_k + \bm{{\rm b}}_{a}), \nonumber 
\end{align}
where $\alpha_{i,k}$ is the weight of $k$-th object $\bm{e}_k$ regarding $i$-th motion;   $\bm{{\rm W}}_a, \bm{{\rm w}}_a$, $\bm{{\rm U}}_a$ and $\bm{{\rm b}}_{a}$ are learnable parameters.

Then, we use a bi-directional LSTM to encode actions:
\begin{align}
    \mathcal{A}= \{\bm{a}_i\}_{i=1}^{T} = \bm{{\rm BiLSTM}}(\{[\bm{m}_i;\bm{m}_{i}^{e}]\}_{i=1}^{T}).
\end{align}
where $\bm{a}_{i} \in \mathbb{R}^{d_{model}}$. The action features $\mathcal{A}$ are later used as inputs for the sentence module and predicate language head.  
Formulated as a fully-connected layer, the predicate language head projects video action to linguistic semantic space:
\begin{equation}
    \bar{\bm{a}}=\bm{{\rm FC}}(\bm{{\rm maxpool}}(\mathcal{A}))
\end{equation}
where $\bar{\bm{a}} \in \mathbb{R}^{d_s}$.

\noindent \textbf{Loss function.} We exploit the \textit{predicate} of   ground-truth captions to supervise our predicate module. Likewise, we use SBERT~\cite{SBert} to encode the \textit{predicate} into embedding $\bm{p} \in \mathbb{R}^{d_s}$. Then, our predicate module is optimized by minimizing the distance between $\bm{p}$ and $\bar{\bm{a}}$:
\begin{align}
     \mathcal{L}_{p} = 1 - \frac{\bm{p} \cdot  \bar{\bm{a}}}{|\bm{p}| \cdot |\bar{\bm{a}}|}.
\end{align}

\subsection{Sentence Module} 
\label{sec:sentence}
Our sentence module is designed to learn global video representations that bridge global visual content and the entire linguistic caption. 
Since a caption comprises \textit{entities}, the \textit{predicate}, and other context information, 
our sentence module takes initial video context features $\mathcal{C}=\{\bm{c}_{i}\}_{i=1}^{T}$, context-related action features $\mathcal{C}^{a}=\{\bm{c}_{i}^{a}\}_{i=1}^{T}$ ,and context-related object features $\mathcal{C}^{e}=\{\bm{c}_{i}^{e}\}_{i=1}^{T}$ as input. 

We obtain $\bm{c}_{i}^{a}$ and $\bm{c}_{i}^{e}$ by attentively summarizing action features $\mathcal{A} = \{\bm{a}_{i}\}_{i=1}^{T}$ and principal object features $\mathcal{E} = \{\bm{e}_{i}\}_{i=1}^{N}$ according to video context $\bm{c}_i$. Here we take $\bm{c}_{i}^{a}$ as an example:
\begin{align}
    \bm{c}_{i}^{a} = \sum_{k=1}^{T} & \beta_{i,k}\bm{a}_{k}, \nonumber \\
     \beta_{i,k} = {\rm exp}(\hat{\beta}_{i,k})& / \sum_{j=1}^{T}{\rm exp}(\hat{\beta}_{i,j}),  \\
     \hat{\beta}_{i,k} = \bm{{\rm w}}_{g}^{\top}{\rm tanh}(\bm{{\rm W}}_{g}&^{\top}\bm{c}_{i}  + \bm{{\rm U}}_{g}^{\top}\bm{a}_k + \bm{{\rm b}}_g), \nonumber
\end{align}
where $\beta_{i,k}$ is the   weight of the $k$-th action regarding the $i$-th context;   $\bm{{\rm w}}_{g}, \bm{{\rm W}}_{g}$, $\bm{{\rm U}}_{g}$ and $\bm{{\rm b}}_g$ are learnable parameters. 

Then, we compute the global video features:
\begin{equation}
    \mathcal{G}= \{\bm{g}_{i}\}_{i=1}^{T} = \bm{{\rm BiLSTM}}(\{[\bm{c}_i;\bm{c}_{i}^{a};\bm{c}_{i}^{e}]\}_{i=1}^{T}), 
\end{equation}
where $\bm{g}_i \in \mathbb{R}^{ d_{model}}$. The global video feature $\mathcal{G}$ is later used as input of the caption generator and the  language head of the sentence module.

The language head of this module is implemented via a fully-connected layer.
It takes $\mathcal{G}$ as input and predicts global video linguistic embedding under supervision of the embedding of captions
\begin{equation}
    \bar{\bm{g}}=\bm{{\rm FC}}(\bm{{\rm maxpool}}(\mathcal{G})),
\end{equation}
where $\bar{\bm{g}} \in \mathbb{R}^{d_s}$.

\noindent \textbf{Loss function.} We exploit the embedding of the whole caption to supervise the sentence module. 
Similarly, the caption embedding $\bm{s} \in \mathbb{R}^{d_s}$ is computed by the pretrained SBERT.
Then we minimize the distance between $\bm{s}$ and $\bar{\bm{g}}$ to optimize our sentence module:
\begin{align}
    \mathcal{L}_{s} = 1 - \frac{\bm{s} \cdot \bar{\bm{g}}}{|\bm{s}| \cdot |\bar{\bm{g}}|}.
\end{align}

\subsection{Description Generation}
\label{sec:decoder}

We employ an LSTM  as our description generator to produce accurate captions in steps. To generate accurate descriptions for videos, our description generator takes all three levels of video representations, their linguistic predictions, and the previous word $w_{t-1}$ as inputs
\begin{equation}
    \bm{h}_{t}^{lang} = \bm{{\rm LSTM}}_{lang}([ \hat{\bm{g}}_{t}^{l}; \hat{\bm{a}}_{t}^{l}; \hat{\bm{e}}_{t}^{l}; \bm{E}(w_{t-1})];\bm{h}_{t-1}^{lang}),
\end{equation}
where $\bm{E}(w_{t-1}) \in \mathbb{R}^{d_w}$ is the word embedding of $w_{t-1}$, and $d_w$ is the embedding size. $\hat{\bm{g}}_{t}^{l}$, $\hat{\bm{a}}_{t}^{l}$, and $\hat{\bm{e}}_{t}^{l}$ are concatenation of visual features and linguistic predictions, \ie, $\hat{\bm{g}}_{t}^{l} = [\bm{g}_{t}^{l};\bar{\bm{g}}], \hat{\bm{a}}_{t}^{l} = [\bm{a}_{t}^{l};\bar{\bm{a}}], \hat{\bm{e}}_{t}^{l} = [\bm{e}_{t}^{l};\bar{\bm{e}}_{t}^{l}]$.
$\bar{\bm{g}}$ and $\bar{\bm{a}}$ are predicted linguistic embeddings of global content and actions as described in Section~\ref{sec:sentence} and Section~\ref{sec:predicate}. 
$\bm{g}_{t}^{l}$, $\bm{a}_{t}^{l}$, $\bm{e}_{t}^{l}$, and $\bar{\bm{e}}_{t}^{l}$ are obtained via attentively summarizing $\mathcal{G}, \mathcal{A}$, $\mathcal{E}$, and entity linguistic prediction $\Bar{\mathcal{E}}$ according to the history hidden state $\bm{h}_{t-1}^{lang}$. These four representations are computed in a similar way. Here we take $\bm{e}_{t}^{l}$ as an example:
\begin{align}
    \bm{e}_{t}^{l} = \sum_{k=1}^{N}&\gamma_{t,k}\bm{e}_{k}, \nonumber \\
    \gamma_{t,k} = {\rm exp}(\hat{\gamma}_{t,k}) &/ \sum_{j=1}^{N}{\rm exp}(\hat{\gamma}_{t, j}),  \\
    \hat{\gamma}_{t,k} = \bm{{\rm w}}_{d}^{\top}\tanh(\bm{{\rm W}}_{d}^{\top}&\bm{h}_{t-1}^{lang} + \bm{{\rm U}}_{d}^{\top}\bm{e}_{k} + \bm{{\rm b}}_{d}), \nonumber 
\end{align}
where $\gamma_{t,k}$ is the weight of object $\bm{e}_k$ at the $t$-th decoding step; and $\bm{{\rm w}}_{d}, \bm{{\rm W}}_{d}$, $\bm{{\rm U}}_{d}$ and $\bm{{\rm b}}_d$ are learnable parameters.

According to the current hidden state of LSTM $\bm{h}_{t}^{lang}$, the probability distribution $\bm{P}_{t}$  over a vocabulary of $D$ words
is computed by a fully-connected layer and the softmax operation
\begin{equation}
    \bm{P}_{t} = {\rm softmax}(\bm{{\rm FC}}(\bm{h}_{t}^{lang})).
\end{equation}

\paragraph{Loss function.} 
Given a video with the ground-truth words $[w_1, \ldots, w_{L_s}]$, where $L_s$ is the caption length, we compute the Cross-Entropy loss to optimize our description generator:
\begin{equation}
    \mathcal{L}_{XE} = -\sum_{t=1}^{L_s}\delta(w_t)^{\top} {\rm log}\ \bm{P}_{t},
\end{equation}
where $\delta(w_t) \in \mathbb{R}^{D}$ is the one-hot encoding of word $w_t$.

However, the long-tailed word distribution problem\footnote{The function words and common words (such as `a', `the', and `man') are far more than the content specific words (such as `lion', `onion', and `bicycle') in the corpus.} in the video captioning corpus likely causes training issues with unbalanced data. 
To alleviate this problem, Zhang \textit{et al.}~\cite{ORG-TRL} propose a plug-in method, where the abundant linguistic knowledge is transferred from a pretrained external language model (ELM) to the description generator.
Given the previous words $w_{<t}$, the probability distribution of ELM at time step $t$ is:
\begin{align}
    \bm{Q}_{t} = {\rm ELM}(w_{<t}|\Theta_{\rm ELM}).
\end{align}
To transfer the knowledge from ELM to the description generator, the KL divergence between $\bm{P}_{t}$ and $\bm{Q}_{t}$ is minimized during training:
\begin{align}
    D_{KL} = -\sum_{d=1}^{D}{Q}_{t}^d\cdot \log\frac{{P}_{t}^d}{{Q}_{t}^d},
\end{align}
where $Q_{t}^d$ and ${P}_{t}^d$ denote the  $d$-th element of $\bm{Q}_{t}$ and $\bm{P}_{t}$.
In this work, we treat $\bm{Q}_{t}$ as ``soft target" of our description generator, and use $D_{KL}$ as $\mathcal{L}_{soft}$.  More details about the ``ELM" and ``soft-target" can be found in ~\cite{ORG-TRL}.

\subsection{Training}

Our model is trained in an end-to-end fashion via optimizing the sum of all losses
\begin{equation}
    \mathcal{L} = {\lambda_{e}\mathcal{L}_{e} + \lambda_{p}\mathcal{L}_{p} + \lambda_{s}\mathcal{L}_{s}} + \mathcal{L}_{XE} + \lambda_{soft}\mathcal{L}_{soft}.
\end{equation}

%% file: section/exper.tex
\section{Experimental Results}
In this section, we evaluate our model on two widely-used datasets: MSVD\cite{DBLP:conf/acl/ChenD11} and MSR-VTT\cite{DBLP:conf/cvpr/XuMYR16},
with four widely-used metrics: BLEU@4\cite{bleu}, METEOR\cite{meteor}, ROUGE-L\cite{rouge}, and CIDEr\cite{cider} (denoted by B@4, M, R, and C, respectively).
We present the results compared to state-of-the-art methods and report ablation studies
on both benchmarks. 

\begin{table*}[htbp]
\footnotesize
  \centering
  \setlength{\tabcolsep}{1.8mm}{
    \begin{tabular}{c|c|ccc|p{3em}ccc|cccc}
    \toprule
    \multirow{2}[2]{*}{Models} & \multirow{2}[2]{*}{Year} &       & Features &       & \multicolumn{4}{c|}{MSVD}     & \multicolumn{4}{c}{MSR-VTT} \\
          &       & Context & Motion & Object & \multicolumn{1}{c}{B@4} & M     & R     & C     & B@4   & M     & R     & C \\
    \midrule
    $\rm{M}^{3}$~\cite{DBLP:conf/cvpr/Wang000T18} & 2018  & VGG & C3D     & -     & \multicolumn{1}{c}{51.8} & 32.5  & -  & -  & 38.1  & 26.6  & -  & - \\
    RecNet~\cite{DBLP:conf/cvpr/Wang00018} & 2018  & Inception-V4 & -     & -     & \multicolumn{1}{c}{52.3} & 34.1  & 69.8  & 80.3  & 39.1  & 26.6  & 59.3  & 42.7 \\
    PickNet~\cite{DBLP:conf/eccv/ChenWZH18} & 2018  & ResNet-152 & -     & -     & \multicolumn{1}{c}{52.3} & 33.3  & 69.6  & 76.5  & 41.3  & 27.7  & 59.8  & 44.1 \\
    MARN~\cite{DBLP:conf/cvpr/PeiZWKST19}  & 2019  & ResNet-101 & C3D   & -     & \multicolumn{1}{c}{48.6} & 35.1  & 71.9  & 92.2  & 40.4  & 28.1  & 60.7  & 47.1 \\
    OA-BTG~\cite{OA-BTG} & 2019  & ResNet-200 & -     & Mask-RCNN & \multicolumn{1}{c}{56.9} & 36.2  & -     & 90.6  & 41.4  & 28.2  & -     & 46.9 \\
    POS-CG~\cite{POS+CG} & 2019  & InceptionResnetV2 & OpticalFlow & -     & \multicolumn{1}{c}{52.5} & 34.1  & 71.3  & 88.7  & 42.0  & 28.2  & 61.6  & 48.7 \\
    MGSA~\cite{DBLP:conf/aaai/ChenJ19}  & 2019  & InceptionResnetV2 & C3D   & -     & \multicolumn{1}{c}{53.4} & 35.0  & -     & 86.7  & 42.4  & 27.6  & -     & 47.5 \\
    GRU-EVE~\cite{GRU-EVE} & 2019  & InceptionResnetV2 & C3D   & YOLO  & \multicolumn{1}{c}{47.9} & 35.0  & 71.5  & 78.1  & 38.3  & 28.4  & 60.7  & 48.1 \\
    STG-KD~\cite{STG-KD} & 2020  & ResNet-101 & I3D   & Faster-RCNN & \multicolumn{1}{c}{52.2} & 36.9  & 73.9  & 93.0  & 40.5  & 28.3  & 60.9  & 47.1 \\
    SAAT~\cite{SAAT}  & 2020  & InceptionResnetV2 & C3D   & Faster-RCNN & \multicolumn{1}{c}{46.5} & 33.5  & 69.4  & 81.0  & 40.5  & 28.2  & 60.9  & 49.1 \\
    ORG-TRL~\cite{ORG-TRL} & 2020  & InceptionResnetV2 & C3D   & Faster-RCNN & \multicolumn{1}{c}{54.3} & 36.4  & 73.9  & 95.2  & \textbf{43.6}  & 28.8  & 62.1  & 50.9 \\
    SGN~\cite{DBLP:conf/aaai/RyuKKY21}  & 2021  & ResNet-101 & C3D & -     & \multicolumn{1}{c}{52.8} & 35.5  & 72.9  & 94.3  & 40.8  & 28.3  & 60.8  & 49.5 \\
    MGRMP~\cite{MGRMP} & 2021 & InceptionResnetV2 & C3D  & - & \multicolumn{1}{c}{55.8} & 36.9 & 74.5 & 98.5 & 41.7 & 28.9 & 62.1 & 51.4 \\
    \midrule
    HMN (ours)  & 2022  & InceptionResnetV2 & C3D   & Faster-RCNN & \multicolumn{1}{c}{\textbf{59.2}} & \textbf{37.7} & \textbf{75.1} & \textbf{104.0} & 43.5 & \textbf{29.0}  & \textbf{62.7}  & \textbf{51.5} \\
    \bottomrule
    \end{tabular}}
  \caption{Comparison with state-of-the-art methods on MSVD and MSR-VTT benchmarks.The best results are shown in bold.}
  \label{total_performance}%
  \vspace{-2mm}
\end{table*}%

\subsection{Datasets} 
\noindent \textbf{MSVD} is a widely-used video captioning dataset composed of 1970 short video clips collected from YouTube. 
Each clip is annotated with up to 40 English sentences.
Similar to existing methods~\cite{SAAT,ORG-TRL,MGRMP,STG-KD}, we take 1200 video clips for training, 100 clips for validation, and 670 clips for testing.

\noindent \textbf{MSR-VTT} is a large-scale dataset for video captioning. 
It is collected from a commercial video website and covers the most diverse visual contents so far.
MSR-VTT contains 10k video clips from 20 categories, such as music, people, and gaming.
Each video clip has 20 English captions, leading to total 28k unique words.
We use the same setup as existing methods~\cite{SAAT,ORG-TRL,MGRMP,STG-KD} for experiments (6513, 497, and 2990 videos for training, validation and testing).

\subsection{Implementation Details}
\noindent \textbf{Feature Extraction and Text Processing.} We uniformly sample  $T=15$ clips for each video.  Each clip consists of 16 consecutive frames. We take the 8-th frame of each clip as keyframe and detected 10 objects for each keyframe.
Similar to most recent methods~\cite{SAAT,GRU-EVE,DBLP:conf/aaai/ChenJ19,ORG-TRL,MGRMP}, we extract context features using the InceptionResNetV2~\cite{IRV2} model
and motion features using the C3D~\cite{DBLP:conf/cvpr/HaraKS18} model.
In addition, we use Faster-RCNN~\cite{Faster-R-CNN} pretrained on Visual Genome~\cite{DBLP:journals/ijcv/KrishnaZGJHKCKL17} to detect object regions.  
All the above-mentioned visual features are projected to a 512-dimension space before being fed into our model. 
Likewise, the dimension of encoded visual features $d_{model}$ is set to 512.

For captions, we first remove punctuations and convert all letters into lowercase. 
Captions are truncated at 20 words and tokenized. 
We exploit off-the-shelf Constituency Parsing tools provided by \textit{AllenNLP}\footnote{https://demo.allennlp.org/constituency-parsing} to extract \textit{nouns} and the \textit{predicate} from the processed caption, where we utilize the \textit{WordNet} to distinguish \textit{entities} from \textit{nouns} as described in Section \ref{sec:entity}.
The embedding size $d_s$ of \textit{entities},  \textit{predicates}, and \textit{sentences} is set to 768 using the off-the-shelf pre-trained SBERT\cite{SBert}.
The size of word embedding $d_w$ is set to 300.

\noindent \textbf{Other details.} We set the number of queries $N$ in Eq.(\ref{eq2}) to 8 for our entity module. Both the encoder and decoder of the entity module have 2 transformer layers for MSVD (3 for MSR-VTT), 8 attention heads, and 512 hidden state size. Empirically, $\lambda_{e}$, $\lambda_{p}$, $\lambda_{s}$, and $\lambda_{soft}$ are set to 0.6, 0.3, 1.0, and 0.5 respectively.
We adopt the Adam~\cite{DBLP:journals/corr/KingmaB14} optimizer with a learning rate of 1e-4 to optimize our model.
The batch size is set to 64, and the training epochs is set to 20. The hidden size of description generator is set to 512.
During testing, we use beam search with size 5 to generate captions. Our entire system is implemented with PyTorch and all experiments are conducted on 1 $\times$ RTX-2080Ti GPU.

\subsection{Comparison to State-of-The-Art Methods}
We evaluate the proposed model against the state-of-the-art methods. 
The main results are presented in 
Table~\ref{total_performance}. 
Our method achieves the best performance on the MSVD benchmark in terms of all the four metrics.
In particular, our method achieves 59.2 of BLUE@4 (with +2.3 improvement) and 104.0 of CIDEr (with +5.5 improvement).
On the MSR-VTT benchmark, our method achieves the best results on 3 out of 4 metrics. 
Note that CIDEr captures human judgment of consensus better than other metrics~\cite{cider}. 
The significant improvements under CIDEr on both benchmarks demonstrate that our model can generate semantically more accurate captions.
Note that
GRU-EVE, STG-KD and ORG-TRL focus on designing complex video encoders to learn video representations, but leave the video-language relevance unexploited. 
SAAT associates the \textit{subject}, \textit{verb}, and \textit{object} with visual information, focusing on local word correspondence. 
However, our model outperforms them by a significant margin (+8.8$\sim$25.9 of CIDEr on MSVD) via jointly modeling the video-language correspondence at three different levels, which indicates the advantage of further exploring the corresponding relationship between the video and sentence.

\subsection{Ablation Study}
\label{sec:ablation}

\paragraph{Effectiveness of Proposed Modules.} 
We evaluate the effectiveness of each module by removing them and retraining the model. 
The results are shown in the first group of Table~\ref{ablation}.
All the proposed modules contribute significantly to the overall performance. 
The entity module contributes the most, followed by the sentence and predicate modules.

\vspace{-4mm}
\paragraph{Effectiveness of Links Between Modules.} In our model, we adapt side connections between modules. 
To evaluate their effectiveness, we cut off such side connections separately. 
We use ``cut e2p'' to denote removing the connection between the entity module and the predicate module, ``cut e2s'' for removing the connection between the entity module and the sentence module,  and ``cut p2s'' for removing the connection between the predicate module and the sentence module.

The results are presented in the second group of Table~\ref{ablation}. 
The performance drops without these side connections, which indicates their contribution to the overall performance. 
Compared to the performance when removing each module completely, the performance is relatively better. 
This indicates that the three modules themselves contribute more than the connection design between them.

\begin{figure*}[t]
    \centering
    \includegraphics[width=.95\linewidth, scale=0.6]{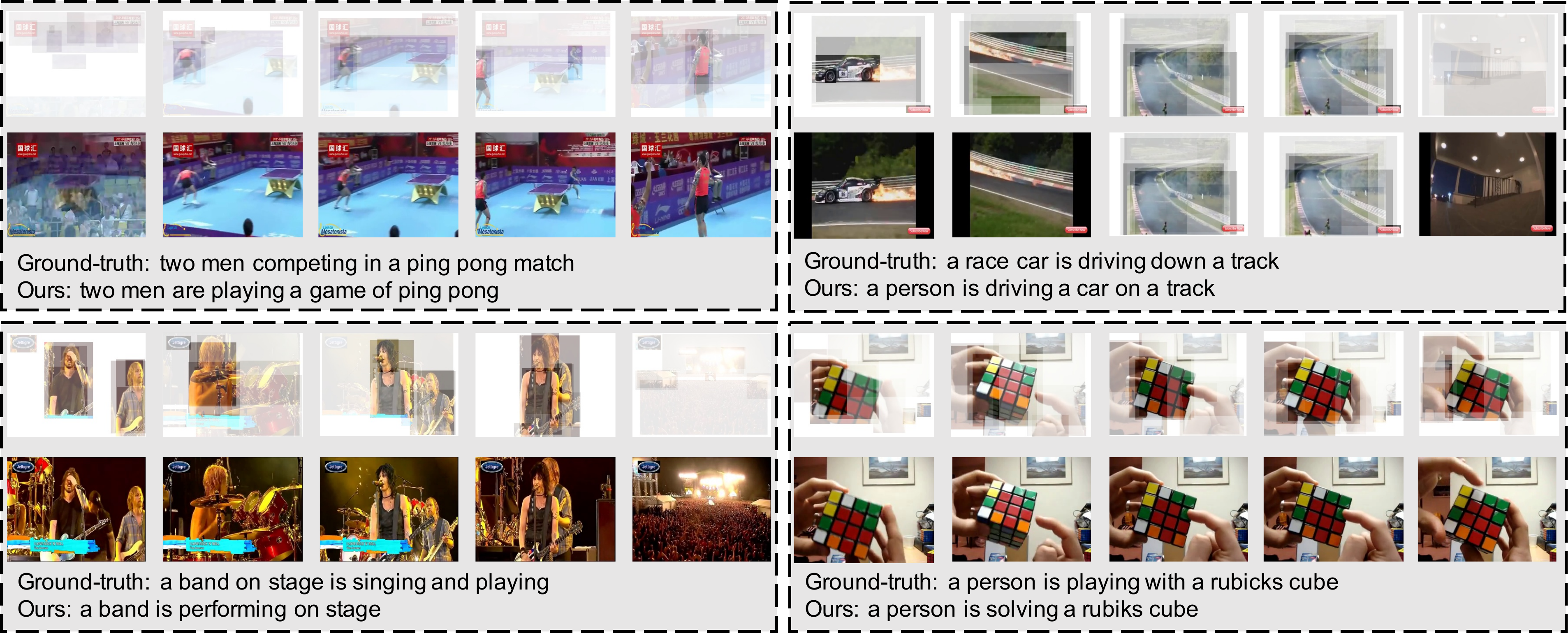}
    \caption{Qualitative results on MSR-VTT. The first row is the principal object regions attended by the our entity module.  A clearer region denotes a higher attention weights.
    The second row shows keyframes for each video.  }
    \label{qualitative}
    \vspace{-2mm}
\end{figure*}

\begin{table}[!tbp]
\footnotesize
  \centering
  \setlength{\tabcolsep}{1.0mm}{
    \begin{tabular}{c|cccc|cccc}
    \toprule
    \multirow{2}[2]{*}{Models} & \multicolumn{4}{c|}{MSVD}     & \multicolumn{4}{c}{MSR-VTT} \\
          & B@4   & M     & R     & C     & B@4   & M     & R     & C \\
    \midrule
    w/o Entity & 51.5     & 34.4     & 71.8     & 88.3     & 40.9     & 27.3     & 60.6     & 46.6 \\
    w/o Predicate & 56.1     & 36.6     & 73.7     & 98.9     & 43.4     & 28.0     & 61.8     & 49.8 \\
    w/o Sentence & 51.8     & 34.8     & 72.2     & 94.8     & 41.5     & 27.5     & 61.0     & 49.4 \\
    \hline
    Cut e2p & 56.9     & 36.7     & 73.9     & 101.3     & 43.3     & 28.0     & 61.7     & 50.4 \\
    Cut e2s & 56.7     & 36.4     & 73.3     & 100.6     & 43.1     & 28.3     & 61.9     & 50.8 \\
    Cut p2s & 56.5     & 37.0     & 73.4     & 100.6     & 42.5    & 28.0     & 61.7     & 50.7 \\
    \hline
    noun supervision & 53.2     & 35.8     & 72.3     & 98.1     & 42.8     & 28.1     & 61.7     & 50.4 \\
    verb supervision & 55.7     & 36.8     & 73.7     & 102.1     & 42.4     & 28.2     & 61.6     & 50.1 \\
    \hline
    all objects & 53.6     & 35.9     & 72.8     & 92.7     & 42.6     & 28.0     & 61.4     & 48.6 \\
    \hline
    w/o v.c.q. & 53.6     & 35.7     & 72.8     & 94.4     & 41.8     & 27.8     & 61.1     & 48.5 \\
    \midrule
    Full model  & \textbf{59.2}  & \textbf{37.7}  & \textbf{75.1}  & \textbf{104.0} & \textbf{43.5}  & \textbf{29.0}  & \textbf{62.7}  & \textbf{51.5} \\ 
    \bottomrule
    \end{tabular}}
    \caption{Ablation studies on MSVD and MSR-VTT.}
  \label{ablation}%
   \vspace{-4mm}
\end{table}%

\begin{figure}[h]
    \centering
    \includegraphics[width=\linewidth, scale=0.7]{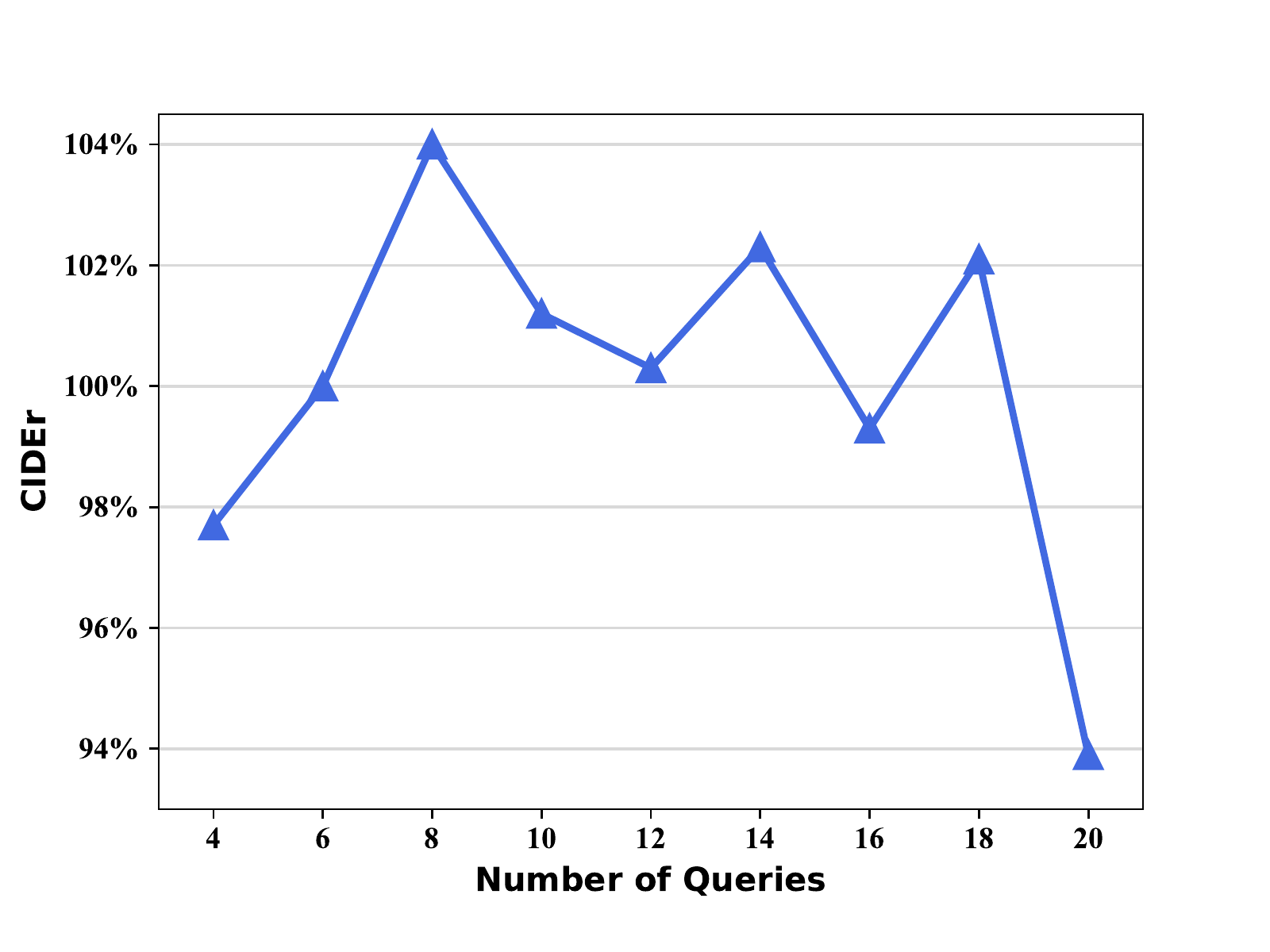}
    \caption{Performance (CIEDr score) comparison under different number of queries.}
    \label{quey_num}
    \vspace{-4mm}
\end{figure}

\vspace{-4mm}
\paragraph{Noun. \textit{v.s.} Entity and Verb. \textit{v.s.} Predicate}
During the training, we supervise the entity and predicate modules with \textit{entities} and the \textit{predicate}, respectively. 
Here we analyze the alternatives by replacing the entity supervision with broader \textit{noun} supervision, and replacing predicate with \textit{verb} supervision.
The results are presented in the third group of Table~\ref{ablation}.
The alternative approaches perform worse than our final model, which indicates \textit{entities} and the \textit{predicate} provide better supervisions for video captioning.

\vspace{-4mm}
\paragraph{Principal Objects \textit{v.s.} All Objects} 
To evaluate the effectiveness of selecting principal objects, we replace our entity module with original embeddings of all objects. 
As row ``all objects'' of Table~\ref{ablation} shows, the model  performance drops significantly.
This is likely caused by the fact that many objects are redundant for video captioning.

\vspace{-4mm}
\paragraph{Video Content Enhanced Object Query} In our entity module, we enhance object queries with video content vectors. 
To verify its effectiveness, we carry out  experiments by removing these vectors. 
The results are presented in the ``w/o v.c.q'' row of Table~\ref{ablation}. 
The model performance drops significantly without video content for queries.

\vspace{-4mm}
\paragraph{Number of Queries} The number of queries $N$ can significantly affect captioning performance as shown in   Figure~\ref{quey_num}. 
Too many or Too few principal object queries can degrade the performance. This is because too many queries may introduce noisy objects and too few queries cannot provide sufficient information. 

\subsection{Qualitative Results}
We show some qualitative results in Figure~\ref{qualitative}.
The first row shows attended objects by our entity module. A clearer region denotes a higher attention weight.
The second row shows the keyframes for reference.
As shown in the figure, our model can generate high-quality captions.
The regions of principal objects can be chosen correctly most of the time, which shows the entity module has the ability to distinguish principal objects from all the detected objects.
We also note that the entity module is able to neglect some redundant frames.
For instance, the example at the top-left shows that our entity module mainly focuses on the two ping-pong players rather than those audiences in the first frame. 
Similarly, in examples at the top-right and bottom-left, our entity module ignores the objects that appeared in the last  frames because their content is not closely related to the video caption. 
Due to space limitation, more qualitative results are placed in supplementary materials.
%

%% file: section/limitation.tex
%
\vspace{-4mm}
\subsection{Limitations}

Similar to existing methods\cite{SAAT,DBLP:conf/cvpr/PanMYLR16,DBLP:conf/aaai/RyuKKY21}, our approach is   suitable for single-action videos. 
However, there exist many multi-action videos in the MSR-VTT dataset. 
For instance, \textit{a man walks in a room and talks to others} has two separate actions, i.e., \textit{walks in a room} and \textit{talks to others}.
Existing methods and the proposed model are not able to handle these complex scenes that require multiple predicates. 
This is also the main reason why the improvement of our model on MSR-VTT is not as significant as that on MSVD.
%

%% file: section/conclu.tex
\vspace{-1mm}
\section{Conclusion}
In this paper, we proposed a hierarchical modular network for video captioning, which bridges video representations and linguistic semantics from three levels: entity level, predicate level, and sentence level. By constructing three modules and associating them with the their linguistic counterparts, we can obtain effective visual representations for captioning.
Furthermore, our entity module is able to identify principal video objects that carry important semantic meanings without processing all mundane ones for generating accurate captions.
Our proposed model has achieved
the state-of-the-art performance on both MSVD and MSR-VTT benchmarks. Extensive experiments and qualitative results have demonstrated the effectiveness of each module. 
\vspace{-4mm}

\noindent \textbf{Acknowledgements:} {\small This work was supported in part by the Italy–China Collaboration Project TALENT under Grant 2018YFE0118400; in part by the National Natural Science Foundation of China under Grant  61836002, 61902092, 61976069, 61872333, 62022083, and 61931008; in part by the Youth Innovation Promotion Association CAS; in part by the Fundamental Research Funds for Central Universities. M.-H. Yang is supported in part by NSF CAREER grant 1149783.
}